\documentclass[conference]{IEEEtran}
\IEEEoverridecommandlockouts

\usepackage{tikz}

\usepackage{filecontents,lipsum}
\usepackage[noadjust]{cite}
\usepackage{amsmath,amssymb,amsfonts}
\usepackage{algorithmic}
\usepackage{graphicx}
\usepackage{textcomp}
\usepackage{url}
\usepackage{booktabs}
\usepackage{hyperref}
\usepackage[table]{xcolor}
\usepackage{tabularx}

\usepackage[numbers]{natbib}

\def\BibTeX{{\rm B\kern-.05em{\sc i\kern-.025em b}%
  \kern-.08em T\kern-.1667em\lower.7ex\hbox{E}\kern-.125emX}}

\begin{document}

\title{Data-Centric AI for Tropical Agricultural Mapping: Challenges, Strategies and Scalable Solutions\\
\thanks{
The authors would like to thank Serrapilheira Institute (grant \#R-2011-37776), Institute of Artificial and Computational Intelligence (IDATA), Brazil National Council for Scientific and Technological Development (CNPq) -grant \#141203/2025-0, National Laboratory of Scientific Computation (LNCC), part of the Ministry of Science, Technology, Innovations and Communications (MCTIC) for providing HPC resources of the SDumont supercomputer, The Brazilian Institute of Geography and Statistics (IBGE), linked to Ministry of Planning and Budget (MPO), and Brazilian Federal Agency for Support and Evaluation of Graduate Education (CAPES) for their financial support for this research.
}
}

\author{\IEEEauthorblockN{Mateus Pinto da Silva, Sabrina P. L. P. Correa, \\Hugo N. Oliveira}
\IEEEauthorblockA{Universidade Federal de Viçosa\\Viçosa, Brazil\\ \{mateus.p.silva, sabrina.correa, hugo.n.oliveira\}@ufv.br}
\and
\IEEEauthorblockN{Ian M. Nunes}
\IEEEauthorblockA{Brazilian Institute of\\Geography and Statistics\\Rio de Janeiro, Brazil\\ian.nunes@ibge.gov.br}
\and
\IEEEauthorblockN{Jefersson A. dos Santos}
\IEEEauthorblockA{University of Sheffield\\Sheffield, \\ United Kingdom\\j.santos@sheffield.ac.uk}
}

\maketitle

\begin{abstract}
Mapping agriculture in tropical areas through remote sensing presents unique challenges, including the lack of high-quality annotated data, the elevated costs of labeling, data variability, and regional generalisation. This paper advocates a Data-Centric Artificial Intelligence (DCAI) perspective and pipeline, emphasizing data quality and curation as key drivers for model robustness and scalability. It reviews and prioritizes techniques such as confident learning, core-set selection, data augmentation, and active learning. The paper highlights the readiness and suitability of 25 distinct strategies in large-scale agricultural mapping pipelines. The tropical context is of high interest, since high cloudiness, diverse crop calendars, and limited datasets limit traditional model-centric approaches. This tutorial outlines practical solutions as a data-centric approach for curating and training AI models better suited to the dynamic realities of tropical agriculture. Finally, we propose a practical pipeline using the 9 most mature and straightforward methods that can be applied to a large-scale tropical agricultural mapping project.
\end{abstract}


\section{Introduction}


\IEEEPARstart{D}{ata-Centric} Artificial Intelligence (DCAI) is an approach to developing AI systems that prioritizes improving the quality of data rather than focusing solely on enhancing models or algorithms \cite{sambasivan2021everyone}. Traditionally, the focus of Machine Learning (ML) development relies on creating and enhancing increasingly sophisticated models (model-centric). 
As emphasized in~\cite{aroyo2022data}, modern models are already powerful enough and, in many cases, the bottleneck lies in the quality of the data.

Some studies have explored the use of Remote Sensing (RS) imagery from a data-centric perspective \cite{roscher2024better}. However, there is still a noticeable lack of research specifically addressing DCAI approaches for agricultural applications \cite{roscher2023data}. The existing literature largely theorizes about how data should be managed, often proposing approaches that are impractical or prohibitively expensive to implement. In light of this, we provide a review of DCAI methodologies that are suitable for large-scale agricultural mapping applications, with a particular focus on their applicability in tropical regions. We highlight the critical role of tropical countries (such as Brazil, Indonesia, and Nigeria) in global agriculture~\cite{oakley2022impactsofagriculture}, and argue that the scarcity of datasets from these regions~\cite{pinto2024tradition} can be effectively addressed through data-centric AI strategies.

Given the scarcity of datasets for tropical regions, we emphasize methods that can be employed for segmentation, noting that parcel delineation—typically addressed through semantic or instance segmentation—is a fundamental step in building large-scale agricultural crop mapping pipelines and datasets \cite{pinto2024tradition}. These methods, however, are not limited to this task and can be readily extended to other applications, such as land use/land cover (LULC) classification and crop mapping.

In this sense, this study aims to present a pipeline for large-scale tropical agricultural mapping using DCAI. We first discuss about mapping in tropical agriculture (Sec.~\ref{sec:tropical_agriculture}), then we present how tradicional assessment is conducted (Sec.~\ref{sec:tradicional-tools}). After that, we present the proposed pipeline (Fig.\ref{fig:machine_learningcycle}) in Sec.~\ref{sec:dcai} and concluding in Sec.~\ref{sec:conclusoes}.
\definecolor{mygray}{gray}{0.85}
\newcommand{\code}[1]{%
  \begingroup\setlength{\fboxsep}{0.7pt}%
  \colorbox{mygray}{\textbf{#1}}%
  \endgroup
}
\newcommand{\refsize}{\scriptsize}

\newcommand{\hfir}{2.4}
\newcommand{\hsec}{1.5}
\newcommand{\hthi}{0.575}
\newcommand{\hfou}{-0.35}
\newcommand{\hfif}{-1.25}
\newcommand{\hsix}{-2.15}
\newcommand{\hsev}{-3.05}
\newcommand{\heig}{-3.9}

\newcommand{\wcreation}{-6.375}
\newcommand{\wcuration}{-2.075}
\newcommand{\wtraining}{2.175}
\newcommand{\wevaluation}{6.4}

\begin{figure*}[ht]
    \centering
    \begin{tikzpicture}[
        tax_node/.style={rectangle, draw=gray!60, fill=gray!15, very thick, minimum size=5mm},align=center]
        
        \node (pic) at (0, 0) {\includegraphics[width=1\linewidth]{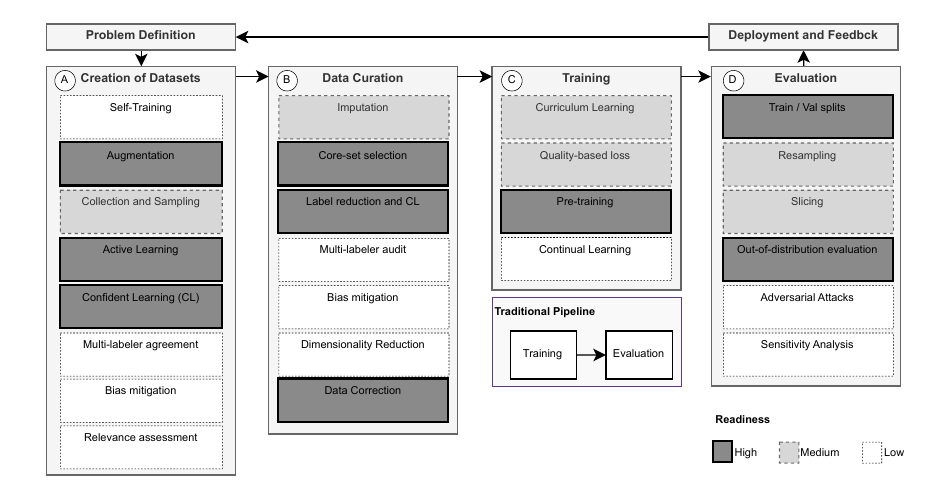}};
        
        \node (creation1) at (\wcreation, \hfir) 
        {\refsize \cite{yang2022st}};
        \node (creation2) at (\wcreation, \hsec) 
        {\refsize \cite{pereira2021chessmix,kim2023bidirectional}};
        \node (creation3) at (\wcreation, \hthi) 
        {\refsize \cite{manas2021seasonal,hong2024spectralgpt}};
        \node (creation4) at (\wcreation, \hfou) 
        {\refsize \cite{lenczner2022dial}};
        \node (creation5) at (\wcreation, \hfif) 
        {\refsize \cite{cheng2019leveraging,corbiere2019addressing,fang2022confident,de2024quality}};
        \node (creation6) at (\wcreation, \hsix) 
        {\refsize \cite{goldstein2021labeling}};
        \node (creation7) at (\wcreation, \hsev) 
        {\refsize \cite{tian2024learning}};

        \node (curation1) at (\wcuration, \hfir) 
        {\refsize \cite{yuan2020self,yuan2022sits,panboonyuen2025satdiff,liu2024diffusion,karwowska2025image}};
        \node (curation2) at (\wcuration, \hsec) 
        {\refsize \cite{nogueira2025core}};
        \node (curation3) at (\wcuration, \hthi) 
        {\refsize \cite{ahn2023fine,cheng2019leveraging,corbiere2019addressing,fang2022confident}};
        \node (curation5) at (\wcuration, \hfif) 
        {\refsize \cite{sun2025mitigating}};
        \node (curation7) at (\wcuration, \hsev) 
        {\refsize \cite{zhang2022vector,lei2024large,cherif2024novel}};

        \node (training1) at (\wtraining, \hfir) 
        {\refsize \cite{zhang2021curriculum,ran2022unsupervised,xi2023multilevel}};
        \node (training2) at (\wtraining, \hsec) 
        {\refsize \cite{dong2021high}};
        \node (training3) at (\wtraining, \hthi) 
        {\refsize \cite{yuan2020self,manas2021seasonal,yuan2022sits,prexl2023multi,XU2024312,pinto2024tradition}};

        \node (evaluation1) at (\wevaluation, \hfir) 
        {\refsize \cite{karasiak2022spatial}};
        \node (evaluation2) at (\wevaluation, \hsec) 
        {\refsize \cite{lyons2018comparison}};
        \node (evaluation3) at (\wevaluation, \hthi) 
        {\refsize \cite{zhang2023adaptive}};
        \node (evaluation4) at (\wevaluation, \hfou) 
        {\refsize \cite{meyer2021predicting}};
        \node (evaluation5) at (\wevaluation, \hfif) 
        {\refsize \cite{tasneem2024improve}};
        \node (evaluation6) at (\wevaluation, \hsix) 
        {\refsize \cite{dixon2025case}};
        
    \end{tikzpicture}
    \caption{Data-Centric AI pipeline, adapted from Roscher \textit{et al.} \cite{roscher2024better}. DCAI approaches and their respective readiness in each step of the ML cycle are presented and contrasted with the traditional supervised ML pipeline. Letters A through D are further discussed in Sections~\ref{sec:dcai_creation}, \ref{sec:dcai_curation}, \ref{sec:dcai_training}, \ref{sec:dcai_evaluation}, respectively. We chose not to include and cite methods with very low levels of readiness due to space constraints.}
    \label{fig:machine_learningcycle}
\end{figure*}
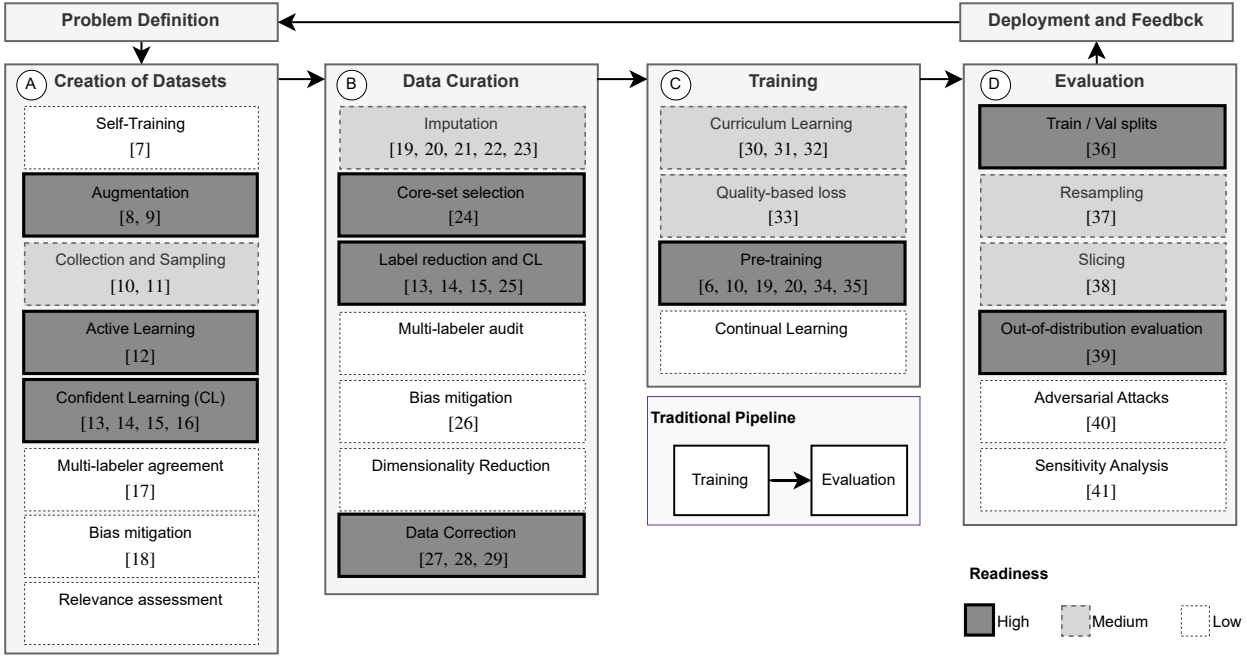



\section{Mapping Tropical Agriculture}
\label{sec:tropical_agriculture}



Agricultural mapping using Deep Learning (DL) techniques applied to remote sensing images has gained prominence in large-scale projects \cite{pinto2024tradition}. One of the main bottlenecks in the process is obtaining high-quality annotations for supervised training. Although the delineation of plots by experts provides a solid basis, imprecise boundaries can introduce noise into the data in semantic segmentation tasks. In addition, uneven frequency of classes may bias the model, requiring rectifying strategies such as resampling and weighted metrics \cite{kussul2017deep}, specially in classification tasks such as Crop Mapping.

Another critical challenge is the generalization of these models, which frequently exhibit a decline in performance when applied to real-world scenarios involving new regions with distinct agroecological characteristics. This problem is exacerbated by the fact that most of the DL agricultural mapping literature has been developed for temperate regionswith strong seasonality, well-defined agricultural calendars, and often annual crops in predictable cycles~\cite{pinto2024tradition}, which do not occur for tropical areas.

In tropical regions, the agricultural calendar is more flexible and crops are more varied in terms of both species and management practices~\cite{moura2020drivers}. This temporal and spatial variability makes it difficult to apply models based on regular time series, widely used in temperate regions \cite{martinez2021fully,sykas2022sentinel}. In addition, high cloudiness for much of the year, especially in the tropical rainforests and Southeast Asia~\cite{feng2024quantifying}, limits the availability of consistent optical imagery. This underscores the need to leverage multi-source data (such as combining radar and optical images) and to develop models that are robust to data gaps.
Synthetic Aperture Radar (SAR) guarantees consistent observations even in adverse weather conditions. Along these lines, Cue \textit{et al.} \cite{cue2019combining} demonstrated that convolutional networks with multitemporal SAR data, combined with prior knowledge about crop dynamics in Brazil, offered gains of up to 8.7\% in F1-score for different crops in Brazil.


\section{Traditional Data Quality Assessment} 
\label{sec:tradicional-tools}

Data quality can be observed as the degree in which a product is adequate to its respective specifications. In specific for geospatial data, there is the ISO 19157:2013 \cite{iso2013spatialdataquality}, which defines elements for spatial data quality\footnote{The elements for spatial data quality are: thematic accuracy, completeness, logical consistency, temporal accuracy, positional accuracy and usability.}. There are two main types of geospatial data: raster images and vector data. Traditional approaches can be generalized as data-agnostic in the sense that the great majority of approaches can be applied to any dataset. Furthermore, this process is usually static, since it does not get revisited by the analysts in the whole pipeline (Fig.~\ref{fig:machine_learningcycle}).


Usually, raster imagery is used for selecting reference samples. RS imagery with sensors capable of detecting electromagnetic radiation either reflected or emitted by the terrestrial surface, with their quality tied to different factors \cite{jensen_remote_2014}. It remains essential to understand the steps involved in image preprocessing and processing to prevent improper use of data. Foundational works on these topics include Jensen \textit{et al.} \cite{jensen_remote_2014} and Schott \textit{et al.} \cite{schott2007remote}.

The other main type of data used in agricultural pipelines is vector data, which typically represents labels as points, lines, or polygons. When processing geospatial data, it is essential to pay careful attention to topological relationships, scale generalization methods, and the use of appropriate map projections \cite{casanova2005banco}. Overlooking these aspects can result in a range of quality issues, including invalid geometries, topological errors, inconsistencies in projection systems, duplicated features, and polygons with erroneous holes. In practice, such problems are often addressed by either manually correcting the data or excluding problematic elements from the dataset.

\section{Data-Centric AI Approaches}
\label{sec:dcai}

In this section, we will delve deeper into the phases of the proposed DCAI pipeline. In each subsection, we describe methods, limitations, challenges, and benefits.

\subsection{Creation of Datasets}
\label{sec:dcai_creation}




Data augmentation methods have proven to be highly relevant regarding Data Creation step. For semantic segmentation, ChessMix \cite{pereira2021chessmix} and Bidirectional Domain Mixup (BDM) \cite{kim2023bidirectional}, for instance, are approaches that aim to promote diversity in the data without significantly compromising its quality. 
Active learning strategies are also promissing approaches by including experts in the annotation process efficiently, although they still require research into the most appropriate heuristics for the agricultural context. The Deep Interactive and Active Learning for semantic segmentation (DIAL) \cite{lenczner2022dial} framework has shown promising results and is one of the few works that propose a complex pipeline for active learning in remote sensing with deep learning. 
Confidence Learning approaches offer mechanisms to identify potentially incorrect labels in already annotated data, and can be applied from pretrained networks; these include Confidence Network \cite{cheng2019leveraging}, ConfidNet \cite{corbiere2019addressing}, and Confident Learning-based Domain Adaptation for HSI Classification (CLDA) \cite{fang2022confident}. Classic confidence measures, such as Normalized Entropy of the boosting (NEP) output and  Best versus Second Best probability (BvSB), can also be useful due to their simplicity and effectiveness in specific scenarios. On the other hand, data collection and sampling strategies are of moderate relevance: although they can contribute to improving the representativeness of data, they generally need to be combined with other approaches to have a significant effect. Examples include Seasonal Contrast \cite{manas2021seasonal}, SpectralGPT \cite{hong2024spectralgpt}, and quality assessment techniques such as in de Simone \textit{et al.} \cite{de2024quality}. Finally, other strategies such as self-training, multi-labeler agreement, bias detection and relevance assessment were considered to have low practical applicability in applications as diverse as tropical agriculture, mainly due to limitations such as the risk of error propagation, lack of human scale, scarcity of robust remote sensing methods or high computational costs \cite{goldstein2021labeling,tian2024learning,yang2022st}.


\subsection{Data Curation}
\label{sec:dcai_curation}



One of the steps of DCAI is Data Curation (Fig~\ref{fig:machine_learningcycle}), which should be seen as a systematic and iterative process, with special attention paid to the diversity, balance, and consistency of the samples, essential factors in guaranteeing the models' ability to generalize \cite{zha_dcai2014}. Data Curation is a fundamental step in the ML cycle, especially in large-scale agricultural mapping scenarios, where data collection is costly and subject to technical limitations such as the presence of clouds or noise in the labels.


Data imputation and inpainting have medium relevance in the agricultural context. These techniques are useful for filling gaps in time series caused by acquisition failures or cloud cover, but they can introduce artifacts that affect segmentation accuracy, especially in critical applications, as such techniques can generate credible but incorrect data. It is therefore necessary to carefully evaluate the cost-benefit of these approaches in a case-by-case basis. Diffusion models have stood out in remote sensing image processing due to their robustness to noise, ability to deal with variability, and training stability. They have been successfully applied to tasks such as image generation, super-resolution, cloud removal, noise reduction, land use classification, change detection, and climate prediction \cite{panboonyuen2025satdiff,liu2024diffusion,karwowska2025image}. Furthermore, recent studies have explored cloud imputation using Transformers with pixel-level time series \cite{yuan2020self} and sequences of small image patches \cite{yuan2022sits} as part of self-supervised pretraining strategies for crop mapping. These approaches will be discussed in greater detail in the Model Training section (see Section~\ref{sec:dcai_training}).


Core-set selection is highly relevant as it reduces redundancy and improves the balance of data sets. Although still little explored specifically for semantic segmentation tasks \cite{nogueira2025core}, these strategies are promising for the agricultural domain. Among the most relevant methods are Label Complexity (LC), Feature Diversity (FD), Feature Activation (FA), Class Balance (CB), and hybrids such as LC/FD and FA/CB. Recent results indicate that some of these approaches outperform the model trained with 100\% of the data, even using significantly fewer samples \cite{nogueira2025core}. In particular, label-based approaches have usually shown the best results on all datasets, with hybrid methods combining diversity and complexity, such as the FA/CB Hybrid, standing out in several situations.

Label noise reduction and confident learning also stand out for their high relevance. These approaches have great potential for improving the quality of labels in noisy data, especially when combined with pretrained models \cite{ahn2023fine}. However, there is still no widely consolidated technique for semantic segmentation tasks, requiring further comparative research. Among the most promising methods are Confidence Network \cite{cheng2019leveraging}, ConfidNet \cite{corbiere2019addressing}, and CLDA \cite{fang2022confident}.

On the other hand, strategies such as multi-labeler error detection, although conceptually interesting, have low practical viability in the context of large-scale agricultural mapping, due to the difficulty of mobilizing and coordinating multiple annotators. Similarly, bias mitigation techniques are relevant from a theoretical point of view, though they still lack robust automatic methods and represent an open line of research \cite{sun2025mitigating}. Also, dimensionality reduction techniques, usually based on feature selection, are of limited use in this context, as they are generally already applied in intermediate stages or outside of neural networks.

Finally, despite being still relatively unexplored, the use of DL approaches for vector data correction has been explored in several applications that particularly benefit from pretrained models. 
UNet-based approaches \cite{zhang2022vector,lei2024large}, proposed for correcting urban and rural road vectors, outperform traditional methods. Another three approaches to vector map alignment \cite{cherif2024novel} proposed: ProximityAlign, with high accuracy in urban areas but high computational cost; alignment based on Optical Flow with DL, efficient and adaptable; and alignment based on epipolar geometry, effective in data-rich contexts, but sensitive to data quality.

\subsection{Model Training}
\label{sec:dcai_training}

Training strategies are essential to ensure robust models, especially in contexts with noisy, unbalanced, or sparsely annotated data. 

Curriculum learning has medium relevance. This strategy organizes the learning process progressively, favoring the robustness and generalization of models by reducing the impact of noise and imbalance. A promising highlight is the work by Zhang \textit{et al.} \cite{zhang2021curriculum}, which proposes a method for transferring models between different regions -- although not yet tested for agricultural applications, it has shown competitive performance in urban benchmarks (Potsdam \cite{weber2021artifive} and Vaihingen \cite{wu2021new}). Other relevant studies include unsupervised and multiscale \cite{ran2022unsupervised,xi2023multilevel} approaches, which reinforce the potential of this technique.

Loss functions weighted by the quality of the data, or quality-based weighted loss, have also been classified with medium relevance. They are particularly useful when labels are uncertain or of poor quality, such as in weak or noisy data. Although interesting, their impact can be reduced if the data has already undergone substantial curation processes, and the computational cost associated with training these models at scale needs to be carefully considered. A representative example is the work of Dong \textit{et al.} \cite{dong2021high}.

Pretraining stood out with high relevance, especially when based on Self-Supervised Learning (SSL). Self-supervision makes it possible to take advantage of large volumes of unannotated data, which is highly advantageous in scenarios with limited annotation resources, such as agricultural mapping. Methods such as Seasonal Contrast \cite{manas2021seasonal} and the multi-task pretraining framework \cite{prexl2023multi} have shown good results, and can be adapted to improve the performance of models in the agricultural domain. Its Inter-modality Simple framework for Contrastive Learning of visual Representations (IaI-SimCLR)~\cite{prexl2023multi} method, for example, is a case of self-supervised application that can be evaluated in this context. Furthermore, pretrained transformer models have also been used on Satellite Image Time Series (SITS), being variations of Bidirectional Encoder Representations from Transformers (BERT). Although the pipeline of pretraining followed by fine-tuning in Crop Mapping is the same, the pretraining methods may vary. Yuan \textit{et al.} \cite{yuan2020self} used regression to impute synthetically generated clouds and cloud shadows, then improved the method using small image patches \cite{yuan2022sits}, while Yijia \textit{et al.} \cite{XU2024312} employed Momentum Contrast (MoCo) pretraining, approximating different views of time series using contrastive learning. Pinto \textit{et al.} \cite{pinto2024tradition} extended this framework to work with non-contrastive Simple Siamese loss, in addition to providing a benchmark comparing the aforementioned methods.

Finally, continual learning was considered to be of low relevance in the current scenario. Although it has applicability in scenarios where models need to be updated continuously, current methods are aimed at very specific domains and have practical limitations for agricultural mapping. Issues such as catastrophic forgetting and high implementation complexity make their use not yet viable in large-scale projects with limited resources.

\subsection{Evaluation}
\label{sec:dcai_evaluation}


Evaluating models in RS contexts presents specific challenges, especially in large-scale agricultural mapping tasks. Data splits for training and validation can significantly impact the perceived quality and generalization capacity of models. 

Robust strategies for organizing training and test data (\emph{Train/test splits}) are highly relevant, since inappropriate geographical divisions can generate inflated performance estimates. Nearby regions tend to have similar patterns, which makes the model's task easier during validation. On the other hand, when the model is tested in geographically distinct areas, performance usually drops, revealing limitations in generalization. Studies such as Karasiak \textit{et al.} \cite{karasiak2022spatial} demonstrate the importance of appropriate spatial division strategies, and tools such as Museo Toolbox have helped to structure these evaluations more reliably.

Spatial resampling techniques, such as spatial k-fold, block cross-validation, and leave-location-out, are considered to be of medium pertinence. They help mitigate spatial bias by creating more demanding and realistic evaluation scenarios. These approaches are especially useful for testing transfer between regions and reducing the impact of class imbalance. The paper by Lyons \textit{et al.} \cite{lyons2018comparison} offers a detailed comparison of these techniques, highlighting the trade-offs between statistical rigor and practicality in the geospatial context.

Slicing was classified as medium direct relevance or applicability. It is a useful technique for evaluating the spatial robustness of models by segmenting the data into sections such as states, biomes, or agro-ecological zones. This division makes it possible to identify systematic error patterns and regional generalization limitations, allowing more refined analyses of the applicability of models in different contexts \cite{zhang2023adaptive}.

Out-of-distribution evaluation is highly relevant, especially for decisions on model reuse or adaptation. Approaches such as the Dissimilarity Index and the Area of Applicability metric~\cite{meyer2021predicting} are promising for quantifying how much a new region differs from those used in training, providing relevant input for risk analysis and decision-making.

On the other hand, adversarial attacks \cite{tasneem2024improve} and sensitivity analyses \cite{dixon2025case} have been classified as having low relevance at present. Although adversarial attacks are an emerging and relevant line of research for model robustness, their practical applicability in agricultural mapping is still limited. Existing sensitivity analysis approaches are generally very specific and need to be adapted to the context of geospatial data and large-scale applications.


\section{Final Remarks}
\label{sec:conclusoes}

To sum up, the usage of DCAI approaches shows potential for large-scale remotely sensed mapping though it still has significant challenges that require careful adaptation of the techniques available in the literature. When analyzing by each topic for DCAI, we can conclude:

\textit{Creation of Datasets:} including imputation strategies \cite{yuan2022sits}, balancing, noise elimination, and data augmentation \cite{pereira2021chessmix}, proved to be as critical, if not more so, than the choice of algorithms themselves. In this context, approaches such as confident learning \cite{fang2022confident}, core-set selection \cite{nogueira2025core}, and semantic data augmentation \cite{pereira2021chessmix} are particularly promising and justify their inclusion in tropical agriculture operational pipelines. Another highlight is the use of multi-temporal information~\cite{pinto2024tradition}, essential for detecting crops in tropical regions. Consideration should also be given to exploiting SAR images \cite{cue2019combining}.

\textit{Data Curation:} has proven critical to ensure robust and reliable performance of mapping models~\cite{kim2023bidirectional, de2024quality}. Moreover, integrating raster and vector data considering geometric correction and topology adjustments should be part of curation process~\cite{cherif2024novel}. Besides, creating an iterative cycle of continuous improvements, identifying and rectifying systematic errors, such as evolutionary learning~\cite{lenczner2022dial}

\textit{Training:} models using SSL pre-training has recently demonstrated significant advantages~\cite{prexl2023multi, pinto2024tradition}. Models pre-trained on unlabeled data from a specific region consistently outperform randomly initialized counterparts, exhibiting greater robustness to regional variability. Moreover, pre-training strategies such as cloud reconstruction~\cite{yuan2020self, yuan2022sits} are particularly beneficial in tropical areas, where persistent cloud cover presents a major challenge.

\textit{Evaluation:} on agricultural mapping applications is arguably one of the most sensitive and least explored topics in architecture. A naive analysis of DL metrics can lead to models performing worse than expected. Spatial division strategies~\cite{karasiak2022spatial} are extremely important to avoid this problem.

Finally, among the techniques studied, those that stand out in terms of their immediate applicability in tropical agriculture mapping, serving as a practical DCAI pipeline is: (i) Confident learning \cite{fang2022confident} and noise detection techniques \cite{ahn2023fine}, due to the possibility of improving the quality of annotated data from pretrained models \cite{yuan2022sits,pinto2024tradition}; (ii) Core-set selection \cite{nogueira2025core}, which can optimize training with less data and less redundancy; (iii) Specialized data augmentation (e.g., ChessMix \cite{pereira2021chessmix}, BDM~\cite{kim2023bidirectional}), adapted to segmentation tasks; (iv) Active learning \cite{lenczner2022dial}, to insert the human expert more efficiently into the curation cycle. These strategies must be carefully combined and adjusted to the tropical agriculture context, seeking an efficient, practical application that takes advantage of the best practices in international literature without ignoring the specific challenges of tropical agriculture. As future work, a benchmark using the pipeline could be done, providing even more practical context for the DCAI area.

\bibliographystyle{ieeetr}
\bibliography{refs}

\end{document}